# Identifying and mitigating bias in algorithms used to manage patients in a pandemic

Yifan Li, B.S.[1]
Garrett Yoon, B.S.[1]
Mustafa Nasir-Moin, A.B.[2]
David Rosenberg, B.S.[1]
Sean Neifert, M.D.[2]
Douglas Kondziolka, M.D.[2,3]
Eric Karl Oermann, M.D.[2,4,5]*

[1] NYU Grossman School of Medicine, New York, NY 10016
[2] Department of Neurosurgery, NYU Langone Health System, New York, NY 10016
[3] Department of Radiation Oncology, NYU Langone Health System, New York NY 10016
[4] Department of Radiology, NYU Langone Health System, New York, NY 10016
[5] Center for Data Science, New York University, New York, NY 10016
*Corresponding Author

Eric Oermann, MD
Assistant Professor of Neurosurgery, Radiology, and Data Science
NYU Langone Health
530 First Ave
Skirball, 8R
New York, NY 10016
eric.oermann@nyulangone.org

**Abstract**

**Introduction** Numerous COVID-19 clinical decision support systems have been developed. However many of these systems do not have the merit for validity due to methodological shortcomings including algorithmic bias.

**Methods** Logistic regression models were created to predict COVID-19 mortality, ventilator status and inpatient status using a real-world dataset consisting of four hospitals in New York City and analyzed for biases against race, gender and age. Simple thresholding adjustments were applied in the training process to establish more equitable models.

**Results** Compared to the naively trained models, the calibrated models showed a 57% decrease in the number of biased trials, while predictive performance, measured by area under the receiver/operating curve (AUC), remained unchanged. After calibration, the average sensitivity of the predictive models increased from 0.527 to 0.955.

**Conclusion** We demonstrate that naively training and deploying machine learning models on real world data for predictive analytics of COVID-19 has a high risk of bias. Simple implemented adjustments or calibrations during model training can lead to substantial and sustained gains in fairness on subsequent deployment.

**Introduction**

Biased algorithms risk worsening disparities and decreasing access in healthcare as modern healthcare relies on algorithms to guide diagnostic and treatment decision making.[1–3] An ongoing "living review" found that only two out of 232 published models are sufficiently reported without bias to merit further validation[4,5], while a recent review of COVID-19 studies found that none of the 2,212 surveyed are of clinical use due to methodological shortcomings.[4,5] This study is the first to assess the risk of naively obtaining a predictive model that is biased by race, age, or sex trained to predict the outcomes of patients suffering from COVID-19 using a real-world dataset consisting of four hospitals in New York City. We subsequently demonstrate how simple adjustments to the training process to account for bias can lead to more equitable models.

**Methods**

We focused on point of care screening for COVID-19 and used overall classification performance demonstrated by area under the receiver-operating curve (AUROC) and sensitivity as key performance metrics. We included all patients who were screened for COVID-19 by nasal swab in hospitals affiliated with the NYU Langone Health System between January 1st 2020 to December 31st 2020. We constructed logistic regression models to predict three clinical cases: requirement for admission, requirement of a ventilator, and mortality. A model was considered *biased* if it failed to have equal sensitivities for each protected feature (see Supplemental Methods for details).[6] Protected feature tested for this study included race, sex, and age over 62 (senior). We trained models naively and with a bias minimization constraint for equal sensitivity (calibrated models). We conducted post-hoc analyses to compare the sensitivity between subgroups in a protected feature using two sample z-tests on a one hundred held-out test sets to simulate how the models would perform in a real-world setting. All analyses used an alpha of 0.05 with Bonferroni adjustment.

**Results**

For the 21,768 patients who screened positive for COVID-19, the median age was between 48 to 52, 48% were female, 32% were over the age of 62, and 56% were a racial minority (non-white). Different subgroups within each sensitive feature have unequal label representation (Supplemental table 1). Naively trained predictive models were found to frequently have a non-zero risk of bias by sex, race, or age (Supplemental table2) as six out of nine analyses showed bias for that particular feature tested. This effect was present for all protected features (Table 1). The naively trained models achieved an average AUC of 0.943, an average sensitivity of 0.527, and an average difference in sensitivity of 0.063 between subgroups. After re-calibrating the models to mitigate bias, the average AUC remained unchanged at 0.943, average sensitivity increased to 0.955, and the average difference in recall between groups of protected class decreased by 75% to 0.016. The recalibration of the models to eliminate bias led to an expected increase in false positives with the magnitude depending upon the underlying distribution of risk (Supplemental Figure 1). After simulating real-world deployment on 100 bootstrapped samples, re-calibration decreased the probability of obtaining a biased algorithm across six out of nine cases and decreased the overall risk of biased predictions by 56%(Table 1).

**Discussion**

We demonstrate that naively training and deploying models on real world data for predictive analytics of COVID-19 has a high risk of being biased by sex, age, or race. While prior studies have suggested a risk of bias[4], and demonstrated it in select cases[1,2], this is the first study to systematically assess the risk of biased model training on real-world data for COVID-19 and provide validation of these concerns. We demonstrate how simple awareness of the problem and easily implemented post-hoc or train time solutions can lead to substantial and sustained gains in fairness on subsequent deploymemnt.[6] There are multiple causes of biased medical algorithms: datasets, modelling decisions, and choice of deployment environment, and addressing these first begins with awareness of the underlying issue. Study limitations include incorporating data only within a single hospital system in New York City and using an easily implemented method compared to more sophisticated pre-existing solutions to reduce bias. The first and most critical safeguard against bias in medical AI models is awareness on the part of physicians and developers, and having a plan to actively screen for bias before deployment. In our increasingly algorithmically driven medical systems, failing to recognize and account for bias has the risk of worsening inequalities.

## Contributors

YL and EKO designed the study. YL and GY conducted the experiments and analysis. YL and EKO drafted the manuscript. All authors provided comments used to finalize the manuscript. All authors approved the final manuscript.

## Acknowledgements

Dr Oermann reported consulting for Google, equity for Artisight Inc, and employment for Merck.

No other disclosures were reported.

No funding was provided for the research

# TABLES
## Table 1: Bootstrap testing of bias correction over trial datasets

| | | Validation Set | | Test Set | |
|---|---|---|---|---|---|
| Predicted Labels | Protected Group | Biased Trials (%) | Successfully Calibrated Trials (%) | Precalibration Biased Trials (%) | Postcalibration Biased Trials (%) |
| Mortality | Sex | 3/100 (3) | 3/3 (100) | 2/3 (66)* | 0/3 (0)* |
| | Race | 4/100 (4) | 4/4 (100) | 1/4 (25)* | 0/4 (0)* |
| | Age | 17/100 (17) | 11/17 (64) | 5/11 (45)* | 0/11 (0)* |
| Ventilator Status | Sex | 1/100 (1) | 1/1 (100) | 0/1 (0) | 0/1 (0) |
| | Race | 1/100 (1) | 1/1 (100) | 1/1 (100) | 1/1 (100) |
| | Age | 3/100 (3) | 3/3 (100) | 0/3 (0) | 0/3 (0) |
| Inpatient Status | Sex | 100/100 (100) | 98/100 (98) | 98/98 (100)* | 9/98 (9)* |
| | Race | 4/100 (4) | 4/4 (100) | 2/4 (50)* | 1/4 (25)* |
| | Age | 100/100 (100) | 93/100 (93) | 93/93 (100)* | 14/19 (15)* |

[*] Post calibration testing have significantly more fair trials

# SUPPLEMENTAL METHODS

We studied all patients who were screened for COVID-19 by nasal swab in hospitals affiliated with the NYU Langone Health System between January 1st 2020 to December 31st 2020. We constructed 100 logistic regression models on all COVID-19 positive patients to predict three clinical cases: requirement for admission, requirement of a ventilator, and mortality. A model was considered *biased* if it failed to satisfy an equal opportunity criteria defined as an equal likelihood of patients from each group in the protected feature being predicted as case positive. The protected features tested for this study included race, gender, and age over 62. For each model, the dataset was split into training (60%), validation (20%) and test (20%). We trained naively and under a bias mitigation constraint. To mitigate bias during training we employed two separate methods. First, we trained models naively, and then after training adjusted the model's predictive threshold to obtain equal sensitivities for each subgroup, while ensuring the model had a greater than 85% sensitivity. Second, we trained penalized logistic regression models using stochastic gradient descent and a penalty for having different sensitivities for each subgroup.

After training, we conducted a post-hoc analysis to compare the model sensitivity between members of each protected feature (e.g., male vs female patients) using two sample z-tests for proportions on the validation set. Then we calibrated the thresholds per subgroup to equalize sensitivities and applied those new subgroup thresholds on the testing set. Lastly, we compared models trained to mitigate bias (post-calibration) versus naively trained models (pre-calibration), on 100 newly-created bootstrapped datasets from the original data to simulate how the models would perform in the real-world pre- and post-bias mitigation. All analyses used an alpha of 0.05 with Bonferroni adjustment for multiple comparisons. We chose to focus on screening models that could guide clinical decision making, and, therefore, our key metrics were overall classification performance demonstrated by area under the receiver-operating curve (AUC) and sensitivity.

## Supplemental Table 1: Characteristics of COVID-19 Positive Patients

| No. of Patients | No. (%) | Gender | | Age | | Race | |
|---|---|---|---|---|---|---|---|
| Type of Patients | Total | Men | Women | Non-senior | Senior | White | Non-white |
| All COVID+ Patients | 21758 | 10549 (52) | 11209 (48) | 14834 (68) | 6924 (32) | 9562 (44) | 12196(56) |
| COVID+ Inpatient Admission | 8851 | 4757 (54) | 4094 (46) | 4280(48) | 4571 (52) | 3819 (43) | 5032(57) |
| COVID+ ICU Patients | 1431 | 958 (67) | 4713 (33) | 552 (39) | 879 (61) | 613 (43) | 818(57) |
| COVID+ Patients Deceased | 1407 | 873 (62) | 534 (38) | 248 (18) | 1159 (82) | 672 (48) | 735(52) |
| COVID+ Patients who Needs Ventilator support | 1180 | 815 (69) | 365 (31) | 445 (38) | 735 (62) | 449 (38) | 731(62) |

**Supplemental table 1:** Different subgroups within the sensitive features (sex, age, and race) have unequal positive label distribution. In general, men require more clinical treatment (e.g., inpatient admission, ventilator support requirement and mortality) than women; seniors require more clinical treatment than non-seniors and non-white patients require more clinical treatment than white patients.

## Supplemental Table 2: Sample Calibration Statistics

| Target Label | Protected Features* | Validation Precalibrated Model | | | | Validation Postcalibrated Model | | | |
|---|---|---|---|---|---|---|---|---|---|
| | | Validation AUC | Validation Recall | p-value (recall) | Absolute Difference of Recalls (95% CI) | Validation AUC | Validation Recall | p-value (recall) | Absolute Difference of Recalls (95% CI) |
| Mortality | Sex | 0.95 | 0.49 | | | 0.95 | 0.96 | | |
| | Male | 0.95 | **0.55** | **0.018** | 0.06 (-0.04,0.15) | 0.95 | 0.96 | 0.873 | -0.00 (-0.04,0.04) |
| | Female | 0.96 | **0.40** | | -0.09 (-0.2,0.2) | 0.95 | 0.96 | | 0.00 (-0.04,0.04) |
| Mortality | Race | 0.95 | 0.49 | | | | | | |
| | Non-white | 0.95 | 0.53 | 0.161* | 0.041 (-0.06,0.14) | | | | |
| | White | 0.95 | 0.44 | | -0.043 (-0.15,0.06) | | | | |
| Mortality | Age | 0.95 | 0.49 | | | 0.95 | 0.95 | | |
| | Non-Senior (<62) | 0.90 | **0.34** | **0.008** | -0.149 (-0.28,-0.02) | 0.90 | 0.90 | 0.096 | -0.042 (-0.12,0.04) |
| | Senior (≥62) | 0.97 | **0.53** | | 0.043 (-0.05,0.13) | 0.97 | 0.96 | | 0.012 (-0.03, 0.05) |
| Ventilator Status | Sex | 0.94 | 0.32 | | | | | | |
| | Male | 0.93 | 0.35 | 0.123* | 0.033 (-0.06, 0.12) | | | | |
| | Female | 0.95 | 0.25 | | -0.064 (-0.17, 0.05) | | | | |
| Ventilator Status | Race | 0.94 | 0.32 | | | | | | |
| | Non-white | 0.94 | 0.31 | 0.810* | -0.006 (-0.10,0.09) | | | | |
| | White | 0.94 | 0.33 | | 0.009 (-0.10,0.12) | | | | |
| Ventilator Status | Age | 0.94 | 0.32 | | | 0.94 | 0.97 | | |
| | Non-Senior (<62) | 0.95 | **0.22** | **0.004** | -0.101 (-0.20,-0.00) | 0.95 | 0.96 | 0.409 | -0.010 (-0.05,0.03) |
| | Senior (≥62) | 0.91 | **0.39** | | 0.073 (-0.03,0.17) | 0.91 | 0.98 | | 0.008 (-0.02,0.04) |
| Inpatient Admission | Sex | 0.94 | 0.77 | | | 0.94 | 1.00 | | |
| | Male | 0.96 | **0.84** | **<0.001** | 0.065 (0.03,0.10) | 0.96 | 1.00 | 0.052 | -0.002 (0,0) |
| | Female | 0.91 | **0.71** | | -0.069 (-0.10,0.03) | 0.91 | 1.00 | | 0.002 (0,0) |
| Inpatient Admission | Race | 0.94 | 0.77 | | | 0.94 | 0.85 | | |
| | Non-white | 0.92 | **0.76** | **0.028** | -0.019 (-0.05,0.01) | 0.92 | 0.84 | 0.828 | -0.01 (-0.04,0.03) |
| | White | 0.95 | **0.80** | | 0.024 (-0.01,0.06) | 0.95 | 0.85 | | 0.001 (-0.03,0.04) |
| Inpatient Admission | Age | 0.94 | 0.77 | | | 0.94 | 1.00 | | |
| | Non-Senior (<62) | 0.90 | **0.65** | **<0.001** | -0.124 (-0.16,-0.09) | 0.90 | 0.99 | 0.457 | 0.030 (0,0.05) |
| | Senior (≥62) | 0.97 | **0.90** | | 0.121 (0.09,0.15) | 0.97 | 1.00 | | 0.033 (0.01,0.06) |

[**BOLD**] values indicate a significance of <0.05
[*] Models were not calibrated if recalls did not significantly differ between subgroups

**Supplemental table 2:** the table shows that six out of nine analyses were biased on the naively trained model. After calibration, the recall differences between subgroups diminished while the AUC remained unchanged.

# SUPPLEMENTAL FIGURES

## Supplemental Figure 1.a

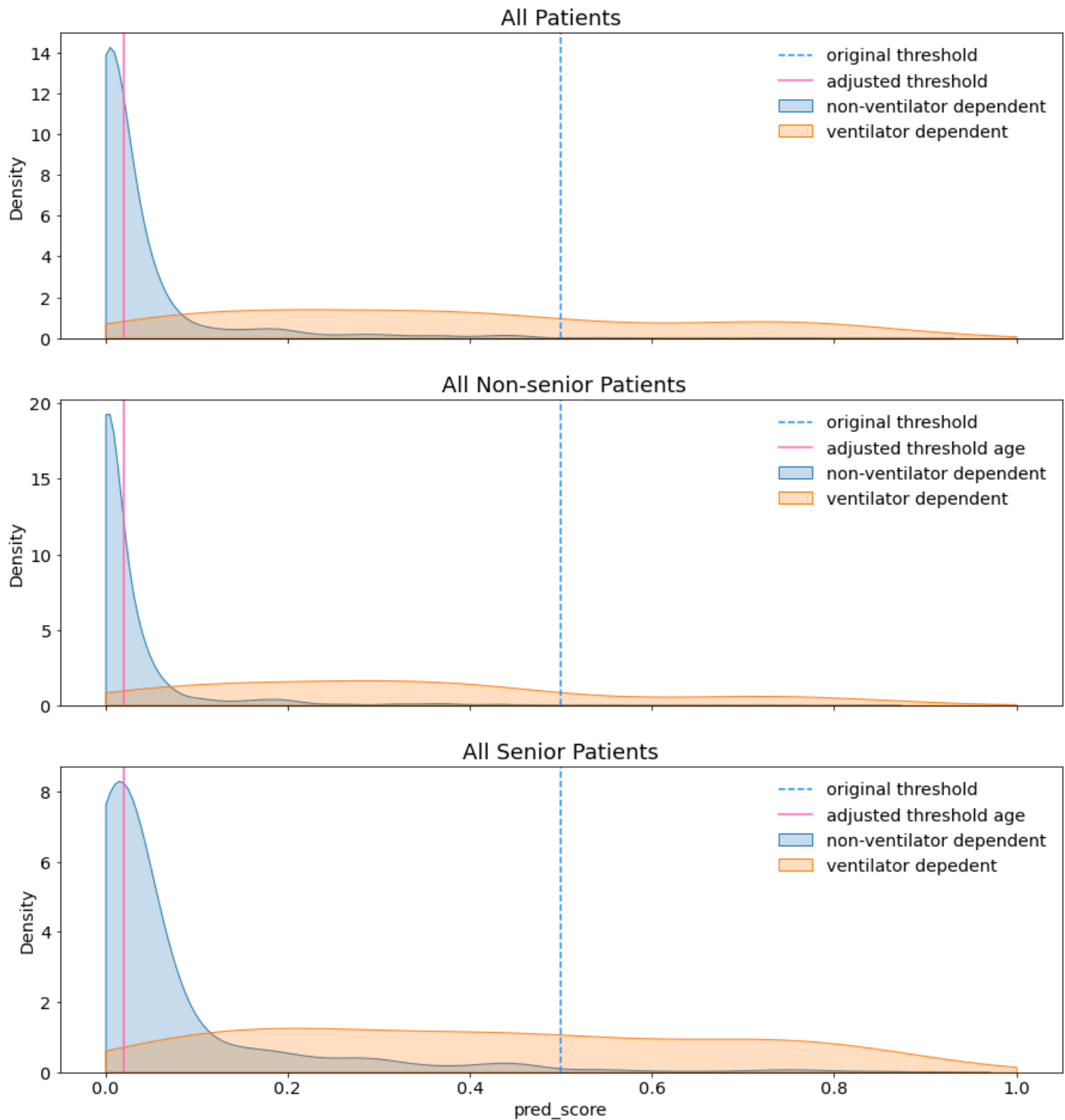

**Supplemental Figure 1.b**

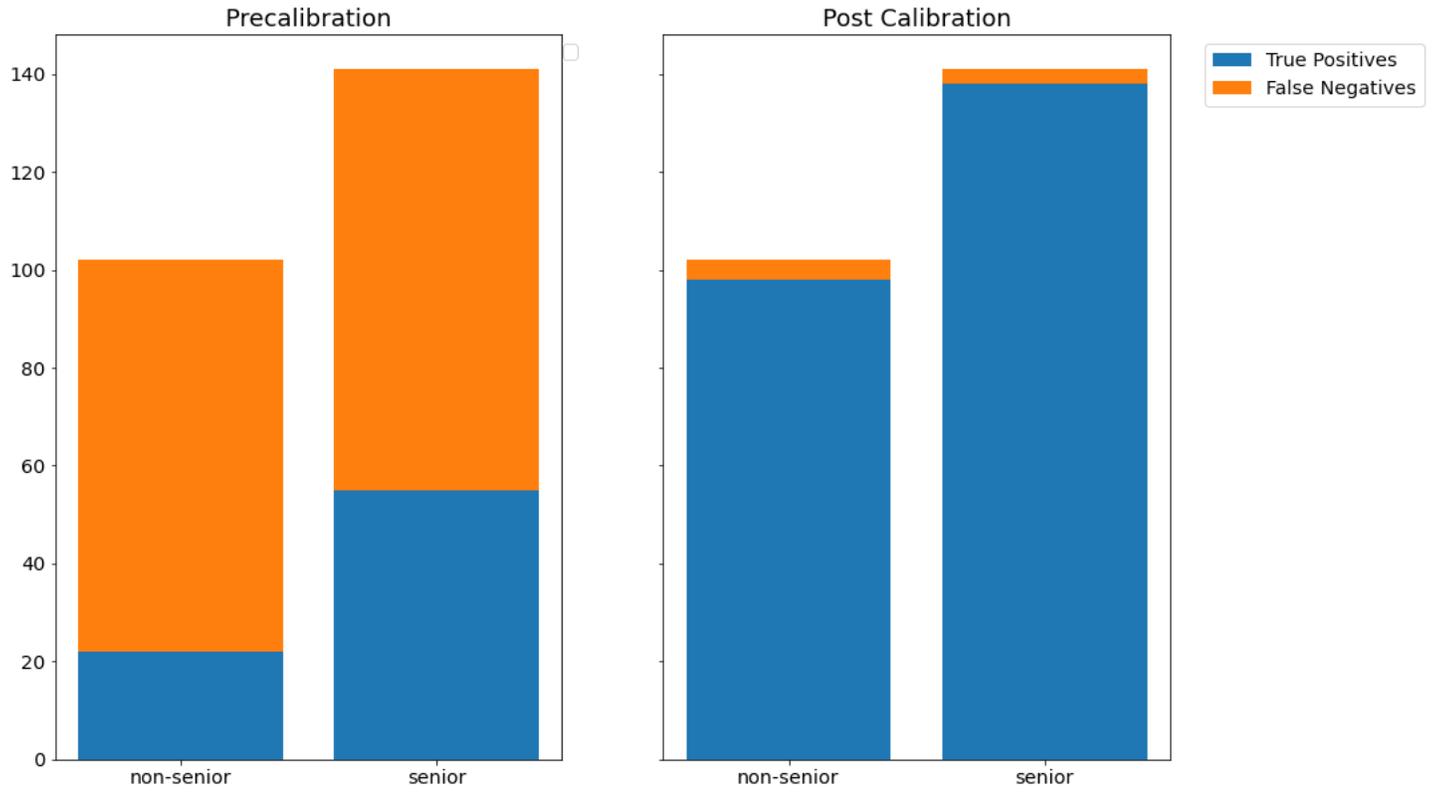

**Supplemental Figure 1.c**

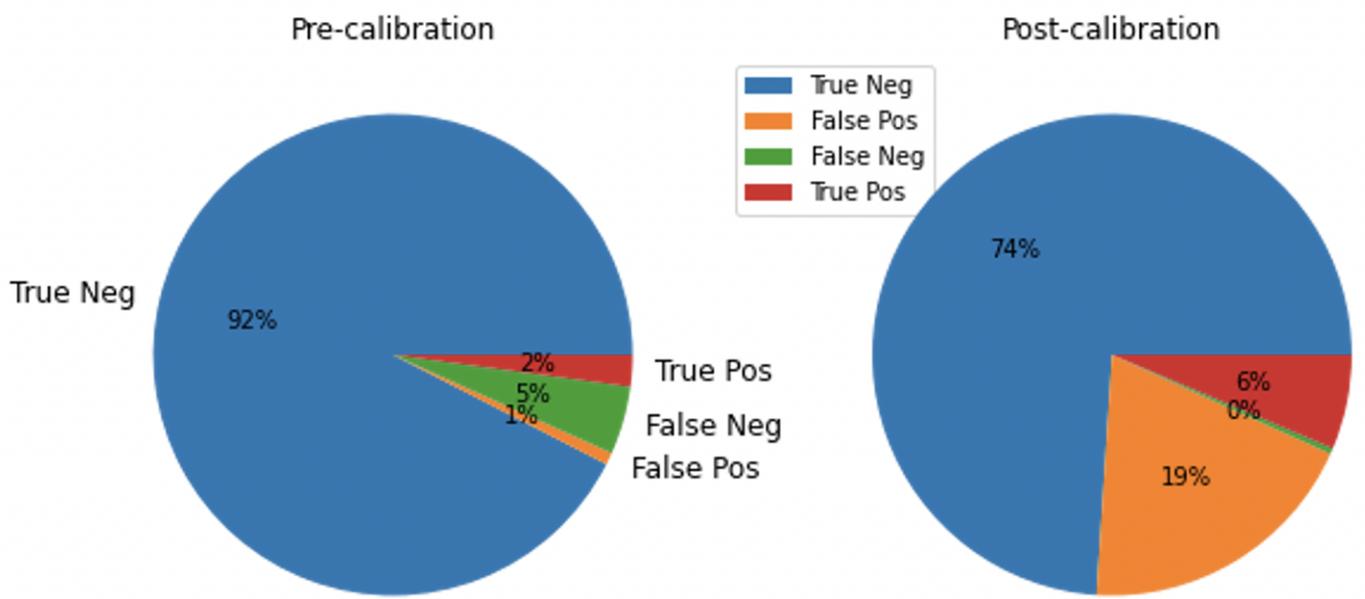

**Supplemental Figure 1: a.** Demonstration of underlying distribution of ventilator requirements based on age for all patients, age less than 62 (non-senior) , and age greater than 62 (senior) demonstrating significantly higher ventilators in the greater than senior group with an original threshold (dotted blue line) chosen to capture this quantity. The bias mitigated threshold (red line) is left shifted to ensure equal recall in the age less than the senior group **b.** Predicted positive versus negative in pre- and post- mitigation age groups showing increase in false positives for age greater than the senior **c.** Calibration largely increased both true positives and false positives and decreased the false negatives. The calibration penalizes accuracy.